\documentclass[conference]{IEEEtran}
\usepackage{hyperref}
\hypersetup{
  colorlinks   = true,    
  urlcolor     = blue,    
  linkcolor    = red,    
  citecolor    = red      
}
\usepackage{amsmath}
\usepackage{graphicx}
\usepackage{todonotes}

\usepackage{color}
\usepackage{algorithm,algorithmic}
\usepackage{url}

\newcommand{\A}{\mathcal{A}}

\renewcommand{\S}{\mathcal{S}}

\begin{document}
\title{Learning body-affordances to simplify action spaces
}
\author{
\IEEEauthorblockN{Nicholas Guttenberg, Martin Biehl, Ryota Kanai}
\IEEEauthorblockA{Araya, Tokyo, Japan \\ 
Email: ngutten@gmail.com, martin@araya.org, kanair@araya.org}}

\maketitle

\begin{abstract}
Controlling embodied agents with many actuated degrees of freedom is a challenging task. 
We propose a method that can discover and interpolate between context dependent high-level actions or \textit{body-affordances}. These provide an abstract, low-dimensional interface indexing high-dimensional and time-extended action policies. Our method is related to recent approaches in the machine learning literature but is conceptually simpler and easier to implement. 
More specifically our method requires the choice of a $n$-dimensional target sensor space that is endowed with a distance metric. The method then learns an also $n$-dimensional embedding of possibly reactive body-affordances that spread as far as possible throughout the target sensor space. 

\end{abstract}

\section{Introduction}
In order for robots to be able to learn to achieve higher-level tasks efficiently they need an awareness of what their embodiment can achieve in the world. This work contributes to the development of automatic ways to endow many degrees of freedom (DOFs) robots with an awareness of their capabilities. In other words, we would like to algorithmically learn the affordances that the body of a robot provides to its controller. 

A fundamental assumption of this work is that not all possible sequences of high-dimensional actuation signals are equally useful. An agent will usually only need a low-dimensional subset of its action space independent of the eventual tasks

that it is faced with. If this is the case it should be possible to learn a low-dimensional embedding of these actions, action-sequences, or policies before going on to more specific applications or higher level cognitive development. For lack of a better term we here call the embedded policies ``body-affordances''. The body-affordances then provide a compressed interface between agent controller and it's body. Awareness of its capabilities corresponds to access to the possible outcomes of these body-affordances. In this work we mostly focus on learning the body-affordances but our method also results in predictions of according outcomes of the body-affordances which could be fed to the controller as well. 

A hint that the above assumption may be justified can be found in biological agents. These often perform dimensional reduction by means of central pattern generators (CPG) \cite{ijspeert_central_2008}. These convert low dimensional signals from higher cognitive levels into time-extended, high-dimensional, coordinated, and reflexive signals that realise appropriate locomotive gaits. These appropriate gaits are either learned during infancy or in other cases (e.g.\ antelopes) possibly hardwired by evolution.
Rather than just providing a reduction of the action sequences the CPGs also directly process sensor inputs to adapt their output. In this sense they provide a lower dimensional choice of closed-loop policies rather than open-loop action sequences. In our language, the set of possible signals to the CPG corresponds to the body-affordances provided to the higher cognitive level.

Other hints that the assumption is justified come from recent advances in reinforcement learning with sparse rewards. There the high-dimensionality of the problem comes less from a high number of actuated DOF and more commonly from the time-extended sequences of actions that have to be invoked in the right combination to get rewards. In the option framework \cite{sutton1999between} time-extended policies called \textit{options} are derived and then added to the choice of possible (elementary) actions. Task independent ways in which the options should be derived are a matter of current research (see Sec. \ref{sec:relw}) and the present work can be seen as proposing such a method as well.

The main intuitions behind our approach are:
\begin{enumerate}
  \item actions or policies can be clustered together to a body-affordance if they lead to the same outcomes
  \item body-affordances should achieve as many different outcomes as possible
  \item small changes in the body-affordance should lead to small changes in the outcome.
\end{enumerate}
How these can be achieved is described in the next section. We note that since we use a predictive network to speed up the learning process our method also results in a way to generate predictions of the outcomes of the body-affordances from the current state. If we feed these predictions back as sensorvalues to a higher level controller together with the body-affordance interface the agent can be seen as being aware of the consequences of its actions. 

\section{Method}
\label{sec:method}
\begin{figure}
 \includegraphics[width=\columnwidth]{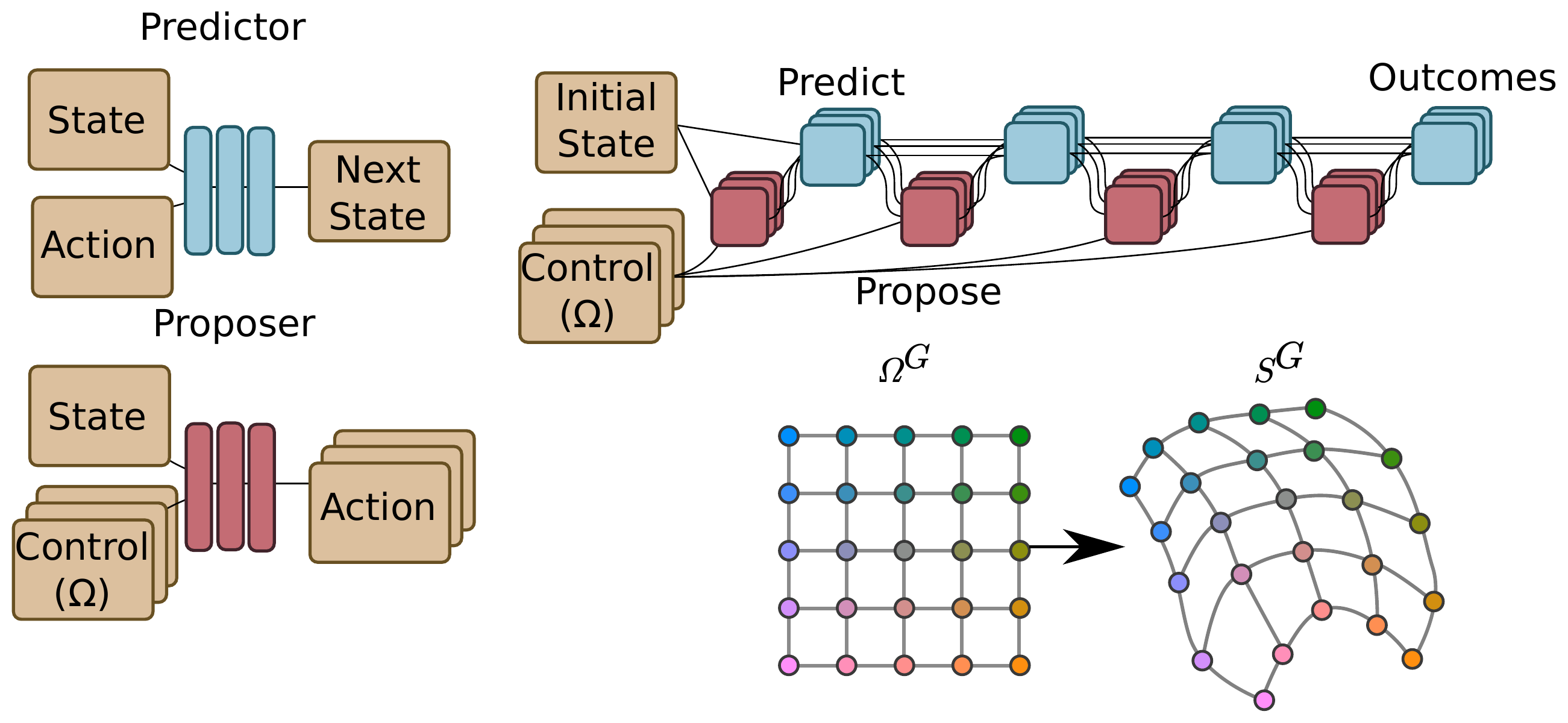}
 \caption{\label{Architecture}Predictor and proposal networks are connected recurrently in order to produce different simulated outcomes. In case of a single action there is only one proposer and one predictor instead of a chain as shown. ``Control'' indicates the space of body-affordances $\Omega$. Bottom right shows a sketch of the body-affordance grid $\Omega^G$ transformed for a given initial state (which is not shown) via interactions of proposer and environment (or predictor) into the outcome grid $S^G$. The proposal network is trained to maximise the differences between the vertices of outcome grid $S^G$. Colour just encodes position in the grid.}
\end{figure}

For our method to work, it is necessary to explicitly choose a continuous standard by which outcomes should be considered distinguishable. This consists of defining a distance $d$ between outcomes in some \textit{target (sub)space} $\S^T \subseteq \S$ of the space of sensorvalues $\S$. Here, we use an explicit time horizon $h$ to determine the point at which a sensor state is considered an outcome, but this could be generalised to variable time horizons by including an explicit stopping action.

It is also necessary to choose the dimensionality and coordinate representation that will be used to construct a control space $\Omega$ to contain the learned body-affordances. In our cases, we pick a finite $n$-dimensional cubic grid $\Omega^G \subset \Omega$ of side length $k$ whose vertices $\omega_{i}\in \Omega^G$ (here $i \in [1,k^n]$) are body-affordances that we use for training. In the case that $\Omega^G$ reliably leads to a regular \textit{grid of outcomes} $\S^G \subset \S^T$ (see bottom right of Fig.~\ref{Architecture}) that pervades $\S^T$ we can then determine a body-affordance that approximately reaches a target point $s^T_{t+h}$ by finding the nearest neighbouring outcomes $s^G_{t+h} \in \S^G$ and then linearly interpolating the values of the according $\omega_i \in \Omega^G$.

First, we train a neural network (the proposer network) $\pi_\theta: \S \times \Omega \rightarrow \A$ to map the current state (or sensor value) $s_t \in \S$ and vertex body-affordances $\omega_i \in \Omega^G$ to actions $a_t \in \A$ (or policies) that lead to maximally different outcomes $s^T_{t+h} \in \S^T$ using the distance on the target space. In order to obtain the outcome $s^T_{t+h}$ for a given time-horizon $h$ the body-affordance $\omega_i$ is fixed such that $\pi_\theta( \;.\; ,\omega_i):\S \rightarrow \A$ corresponds to a fixed reactive policy that interacts with the environment for $h$ timesteps. The simplest way to force the outcomes to be different and thus the body-affordances to pervade the target space is to maximise the minimum distance between them, but more sophisticated methods are possible 
(see e.g.\ Sec. \ref{sec:ex}).

In general the environment is a black box which we cannot explicitly differentiate so that learning will be slow. However, we can train a forward predictive model to emulate the environment, 
and then optimise the proposer with respect to that.
So in addition to the proposer $\pi_\theta$ we train a predictor network $\gamma_\phi:\S \times \A \rightarrow \S$ which maps a given sensor value $s_t \in \S$ (not $\S^T$) and an action $a_t\in A$ to a next estimated sensor value $\hat{s}_{t+1}$. We can then chain proposer and predictor together starting from a given initial state (or sensor value) $s_t$ to produce estimated outcomes $\hat{s}^T_{t+h}\in \S^T$ (see Fig.~\ref{Architecture})\footnote{In principle this method can be augmented by allowing the predictor and proposer networks to also pass themselves latent variables. 
}. 

Note that since both networks need to understand the sensorvalues $s_t \in \S$, it is convenient to allow both networks to share a few layers purely for processing sensor inputs, before fusing with the proposed action (for the predictor) or the body-affordance (for the proposer).

The predictor can be trained via supervised learning, minimising the mean squared error between the predicted outcome and the actual outcome. If we have sufficient data in the form of triplets $(s_t,a_t,s_{t+1})$ about a robot/agent both proposer $\pi_\theta$ and predictor $\gamma_\phi$ can be trained in an offline manner. It is also possible to add data continuously during training. In all cases it is generally necessary to perform several gradient descent steps per observation in order to extract all of the useful information contained in that example. 

In practice, it can be useful to use a partially trained proposal network to generate the action policies which the predictor learns to predict, since that will cause the predictor to become more specialised towards what the proposal network is actually trying to do. To this end, training can involve a cycle consisting of:

\begin{algorithm}
\caption{Training procedure}
\begin{algorithmic}[1]
 \STATE Collect triplets $(s_t,a_t,s_{t+1})$ based on proposal network plus random variation;
 \STATE Add to experience dataset;
 \STATE Train predictor $\gamma_\phi$ to convergence on the entire dataset;
 \STATE Train proposer $\pi_\theta$ to convergence on the predictor;
 \STATE Repeat 
\end{algorithmic}
\end{algorithm}

\section{Related work}
\label{sec:relw}

While the three intuitions mentioned above are reasonably straightforward, they are also closely related to the more theoretically principled approach of selecting the body-affordances so that they maximise empowerment \cite{klyubin_empowerment_2005}. Empowerment is the channel capacity from the actions (here the body-affordances) to future sensor values. It is maximised if the variability (more precisely the entropy) of the future sensors is high and the body-affordances can reliably determine them (conditional entropy of the sensor values with respect to the body-affordances is low). Empowerment has been used more directly in order to derive body-affordances (under a different name) in \cite{gregor_variational_2016}. 

Their work ultimately uses a method where options are defined implicitly with respect to outcomes. However, they comment that an intermediate learned hidden layer representation could be used to obtain a lower-dimensional option/affordance space. We tried several implementations based on this idea, but found that problems could arise from the fact that the training data only mapped to some subset of the option space. Points outside of the subset would generally map to nonsensical action policies.

Our approach attempts to address this by fixing the structure of the latent space to be completely covered by points that the network must make separate in the outcomes. As a consequence, we lose the ability to directly estimate the information about the actions contained in the final sensorvalues (which would be used to calculate empowerment), but instead attempt to maximise empowerment by choosing actions for the grid of body-affordances $\Omega^G$ which result in maximally separated outcomes. Under the assumption that the source of unreliability in achieving an outcome is Gaussian noise with the same variance for each body-affordance our method also maximises empowerment.

Earlier work that also uses empowerment to find options are \cite{anthony_impoverished_2009,anthony_general_2014}. However, this work only treats finite action spaces and it is unclear whether it scales to continuous spaces. 
Another interesting and successful approach to dimensional reduction \cite{thomas_independently_2017} focuses more on finding policies that independently control features in the environment than maximising control itself. A combination with a control maximising method would be interesting further work.

\section{Experiments}
\label{sec:ex}
We use Bullet physics engine \cite{bullet} as an environment simulator, and implement our networks in pytorch \cite{pytorch}. Code for our experiments is available at \url{https://github.com/arayabrain/AffordanceMapping}.

\subsection{Reaching task}

We first consider the case in which the agent is learning to take single, complex actions --- for example, it has a body with many DOF and we wish to represent that large set of DOF with a much lower-dimensional body-affordance space $\Omega$, but we aren't concerned yet with sequences of multiple actions over time.

The task we use is to control the reaching behaviour of a segmented armature. The armature is composed of 9 cylindrical segments connected in sequence by 8 hinge joints, which can rotate only within a limited range of angles. The armature can reach points within a roughly hemispherical shell around its base of radius 4 units. In addition, the environment includes a variety of randomly positioned spherical and cubical obstacles. A 24 pixel resolution depth camera is suspended above the armature and provides sensory information about the environment to the predictor and proposer networks. 

\begin{figure}
 \includegraphics[width=\columnwidth]{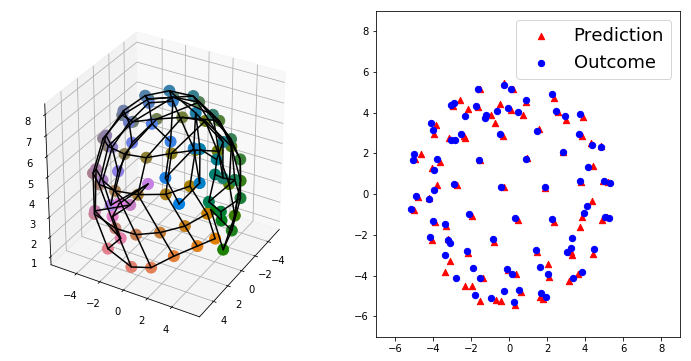}
 \caption{\label{ReacherMesh}The left figure (a) shows the outcome grid $\S^G$ of the robot arm. Points are coloured according to the values of the body-affordance grid coordinates $\omega_i\in \Omega^G$ (compare bottom right of Fig.~\ref{Architecture}). The right figure (b) shows the top down view of these points according to the actual outcome (blue) and the predicted outcome (red). }
\end{figure}

The target space $\S^T$ is the space of reachable positions of the tip of the armature. While this is technically three-dimensional only an approximately two-dimensional sub-manifold can be reached. We choose the two-dimensional grid $\Omega^G$ with $k=9$ such that it covers all of $\Omega = [-1,1] \times [-1,1]$. To force differences between the outcomes $s^G_{t+1}$ of the $\omega_i \in \Omega^G$ we maximise their minimum distance. 
In addition, we add a term to the loss function minimizing the distance between neighboring pairs of grid points, which helps ensure the smoothness of the grid.

\begin{figure}
 \includegraphics[width=\columnwidth]{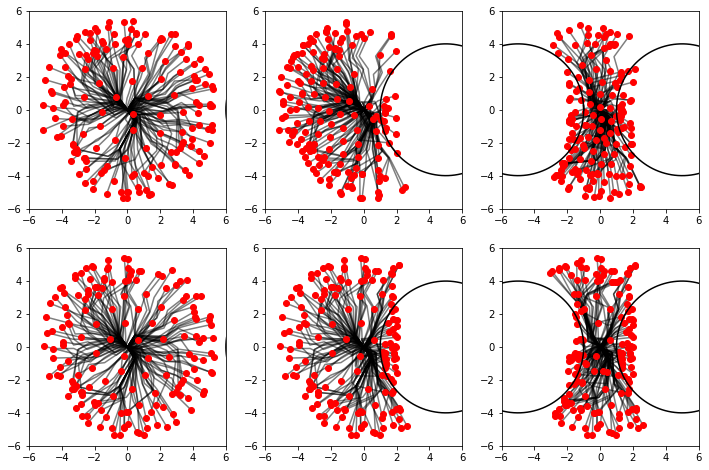}
 \caption{\label{ReacherObstacle}
 Top: Outcome grid $\S^G$ for three different environments with camera seeing the obstacles. Bottom: Outcome grid obtained with camera image fixed to see no obstacles. When the proposer is shown the correct environment, it redistributes the proposed actions accordingly to make better use of the body-affordance space.
 }
\end{figure}

In practice we found it necessary to use very small learning rates for the proposer to maintain the smoothness of the target space with respect to the body-affordance space.

We iteratively generated a dataset according to the above algorithm, consisting of 200000 random environments and joint angles and 90000 environments and joint angles taken from subsequent partially-trained proposal networks. The final predictor was trained from scratch for 150 epochs on the full dataset, trying to minimise the mean-squared error of the final position of each segment of the armature, achieving a final mean squared error of $0.076$ on a held-out testing set of $5000$ environments and angles. Given that the reachable space is a hemisphere of radius 4, this corresponds to about a 4\% positioning error in predicting where the tip will end up. An example of actual versus predicted points is shown in Fig.~\ref{ReacherMesh}b.

In Fig.~\ref{ReacherMesh}a, we show the outcome grid $\S^G$ in the absence of any environmental obstacles. Since our body-affordance space $\Omega$ has a planar geometry, there are some areas where the grid does not extend over the entirety of the reachable hemisphere, but in general those points are fairly close to the opposing side of the outcome grid. As a result, the interpolation scheme mentioned in Sec. \ref{sec:method} could be made to work. The outcome grid $\S^G$ gives an idea of what the reacher arm is able to reach, while affording this to a controller via a 2 DOF interface rather than an 8 DOF one. 

When we add obstacles, the proposer is still able to maintain a fairly uniform coverage of the reachable space. In Fig.~\ref{ReacherObstacle}, we show an overlay of the different configurations of the reacher in the presence of different obstacles corresponding to points on the proposed outcome grid (top row). Compared to the case of simply transplanting the body-affordances from the obstacle-free environment (bottom row), the proposed outcome grid is much more uniform even when the reacher goes from being fairly unconstrained to being so constrained that it loses the ability to bend at the trunk. 

\subsection{Closed loop control: hexapod}

We now consider the case of learning multiple-action body-affordances of a hexapod robot. The hexapod has three hinge joints per leg, each of which is controlled via a target angle. Since we are concerned with locomotion specifically, we provide a sinusoidal clock signal to the model and ask the actions to determine the phase angle and amplitude for how that clock signal is applied at each joint. The robot has its centre of mass position and orientation as well as the joint angles as sensor inputs, and provides 36 actions at 5 time points during a run. We reduce this 180 dimensional policy space down to a 2 DOF body-affordance space $\Omega$, using the final in-plane displacement of the centre of mass as the target space $\S^T$.

We find that iterative training predictor and proposer is more important to keep the predictor accurate compared to the reacher. We also find that body-affordance space can be very discontinuous due to collisions of the feet with the ground. 
To increase robustness, we add noise to the sensor and action values when training the predictor, and also ask the predictor to assess its own uncertainty by outputting the most likely parameters of a Gaussian distribution modelling the outcome rather than just a single estimated point. We then add a regularising term to the proposer's loss function equal to $-\alpha \log(\langle \sigma \rangle)$ where $\alpha = 0.01$ is the strength of the regularisation and $\langle \sigma \rangle$ is the mean standard deviation over all predicted variables. This encourages the proposer to avoid high uncertainty points while trying to spread out the outcome grid.

\begin{figure}
  \includegraphics[width=\columnwidth]{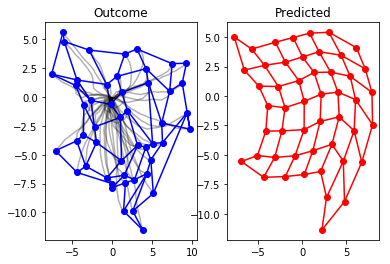}
\caption{\label{Hexapod}Left: Outcome grid $\S^G$ of centre of mass positions of the hexapod. Black lines show the trajectories of the centre of mass. Right: Predicted outcome grid.}
\end{figure}

The resulting outcome grid and corresponding centre of mass motions are shown in Fig.~\ref{Hexapod}. We observe that the hexapod has learned to move its centre of mass quite some distance away from its starting point. The robot is 2 units in radius, but after 10 cycles of its gait it has moved up to 10 units away in the most extreme cases. The body-affordances enable it to reach a number of different final positions, so that the hexapod not only learns to walk but can also dynamically change the target to which it is walking. The outcome grid is noisier than in the case of the reacher (though significantly less noisy than when we train the predictor without asking for uncertainty estimates), and there is some overall rotation which seems to be due to the orientation of the centre of mass drifting during the actual policy playout.

\section{Conclusion}

We proposed and tested a method for dimensional reduction of action spaces that learns closed loop controllers that provide a low-dimensional interface to higher-level control. We were able to construct 2 DOF interfaces for both, 8 DOFs of a robot arm and a 180 DOF time-extended hexapod action space. In the case of the hexapod, the learned control space extended to the discovery of locomotive gaits, allowing the robot to reach different points on the plane. Furthermore, the control spaces produced by this method tend to be smooth and interpolatable.

In the future, we would like to evaluate the effect that using these intermediate controllers has on the rate of reinforcement learning at higher levels, to directly test whether or not this addresses issues of sparse rewards. Furthermore, we would like to see if it is possible to relax the requirement to impose a distance metric on the outcome space while retaining the guarantee of complete coverage over the option space.

\bibliography{affordance.bib}

\end{document}